\documentclass[runningheads]{llncs}
\usepackage[T1]{fontenc}
\usepackage{graphicx}
\usepackage{makecell}
\usepackage{multirow}
\usepackage{diagbox}
\begin{document}
%
\title{Age-Adaptive Handwriting Reconstruction from an IMU-Based Digital Pen through Shared Representations and Domain-Specific Heads}
\titlerunning{Age-Adaptive Handwriting Reconstruction from a Digital Pen}
%
\author{
Florent Imbert\inst{1} \and
Yann Soullard\inst{2} \and 
Eric Anquetil\inst{2} \and
Hui Han\inst{1}
} 

\authorrunning{F. Imbert et al.}

\institute{
Machine Learning Group, Luleå University of Technology, Luleå, Sweden 
\email{florent.imbert@associated.ltu.se, hui.han@ltu.se}
\and
Univ Rennes, CNRS, IRISA - UMR 6074, Rennes, France
\email{yann.soullard@irisa.fr, eric.anquetil@irisa.fr}
}

\maketitle              
\begin{abstract}
Digital pens are widely used to capture handwriting on digital devices, enabling precise trace recording and enhancing human-computer interaction. However, most are bundled with tablets and lack cross-brand compatibility. Recent digital pens equipped with kinematic sensors have emerged, designed for use on any surface. This especially opens significant potential for supporting handwriting acquisition in classrooms. 
Handwriting reconstruction from such an IMU-equipped pen poses a challenge due to the significant variability in sensor signals between adults and children. Even when producing visually similar traces, variations in writing dynamics, motor control, pen holding, and user confidence introduce substantial discrepancies in the captured signals. Additionally, the high variability in children's handwriting requires collecting large amounts of data, which is not feasible to implement in a school environment at scale. 
Furthermore, models trained exclusively on adult data fail to generalize to children's handwriting, and conversely, models trained on children's data perform poorly on adult writers. This cross-population degradation highlights the need for a unified model that can be deployed directly on the pen, without any user-specific adaptation. 
To address this issue, we propose a cross-domain learning, using an original neural network architecture based on a Temporal Convolutional Network and multiple prediction heads. The model is designed to be robust across age groups by leveraging shared features while effectively handling variability induced by differences in graphomotor development. This approach aims to improve handwriting trace reconstruction from sensor data, where each domain benefits from additional data provided by the other domain.

\end{abstract}
%
%
%


\section{Introduction}

Digital pens have been increasingly developed, offering various uses depending on their capabilities, such as note-taking, handwriting learning, or drawing \cite{signer2024pen}. Initially, electronic styluses became widespread through their association with tablets. These proprietary styluses rely on digital ink and are distributed with tablets, without interoperability across brands. Other digital pens enable writing on paper while keeping a digital trace. As the well-known Anoto Digital pens, these pens integrate a camera (an optical sensor, often an infrared camera) to capture the writing trace and require the use of special glyph-patterned papers (dots or dashed), thus limiting the tool's compatibility. Few digital pens, such as the pen from Lunardini et al.~\cite{lunardini2020smart} and the STABILO Digipen~\cite{KIHT2,KIHT1}, are equipped with Inertial Measurement Unit (IMU) sensors, and do not integrate a camera. The objective is to free oneself from the constraint of a special paper and can also be used on a tablet with a suitable ink. In academic settings, handwriting has been shown to offer significant advantages over typing on a keyboard for children’s learning and knowledge acquisition~\cite{ihara2021advantage,marano2025neuroscience}. In addition, IMU-equipped pens offer significant potential for enhancing handwriting learning with adapted pedagogical tools. 

The STABILO Digipen has been designed with the goal to use it on any surface. It registers accelerations and rotation rates in three axes and the force on the writing tip. A data acquisition protocol was implemented to collect handwriting data with the Digipen~\cite{harbaum2024kiht}, along with the associated ground truth, i.e., the traces recorded by a tablet
, enabling training a neural network. Although IMU are widely used in motion tracking due to their low cost and portability, the major disadvantage of using such IMU-based pens is that the signals from these sensors are relative movement information, often noisy, which makes the task of reconstructing the handwriting complex. This is particularly critical for handwriting applications, where precise trajectory recovery is essential, especially in e-learning scenarios requiring fine-grained feedback. Moreover, children’s handwriting signals exhibit far greater complexity than those of adults, owing to variable graphomotor skills that leads to variations in speed, fluidity, and regularity based on writing proficiency. Consequently, building a robust model for reconstructing children’s handwriting from IMU sensor data is highly challenging, as it demands extensive child-specific data acquisition to account for the wide variability in their writing, which is not easy to get in educational environments. In addition, models trained exclusively on adult data perform poorly when evaluated on children, and models trained on children's data fail to generalize to adult writers~\cite{ADAPDA}. 
A natural question then arises as to whether a model trained on adult data could be leveraged to handle children's handwriting, and vice versa. This motivates the design of a single unified model capable of handling both populations, which could be embedded directly on the pen without requiring any prior knowledge of the writer's age.


In this work, we address the problem of reconstructing handwriting trajectories from both adults and children data using the Digipen. We propose a cross-domain learning based on an original neural network architecture designed to handle handwriting produced across age groups. Our neural network leverages shared features while effectively handling variability induced by differences in graphomotor development. The goal of this work is twofold: first, to leverage data from both adults and children to improve handwriting reconstruction tasks for each age group, thereby mitigating the scarcity and high variability of children's data, which is difficult to acquire; second, to use a single model to handle both adult and child handwriting, moving toward a universal system.  

\section{Handwriting reconstruction using the Digipen}

Handwriting reconstruction has varying levels of complexity depending on the acquisition surface and the target population. Especially, the friction on the writing surface and the fluidity of the writing motion induced by the expertise in graphomotor skills are all factors that impact the quality of the handwriting reconstruction. Table~\ref{tab:comparison} provides an overview of existing approaches along these two axes, and assigns a difficulty level to each configuration. 

Reconstructing handwriting from a tablet for adult writers is considered the simplest setting
, as the device enables smooth pen gliding with minimal drag and friction, resulting in less-noisy signals. In addition, large annotated datasets are readily available for this target population and easier to acquire. Moving to paper-based surface increases the difficulty
, as the friction between the pen and the paper surface introduces significant noise in the captured signal, degrading the quality of the acquired trajectories. 
Handling children's handwriting on tablet introduces additional challenges
, due to the high intra-class variability inherent to developing the act of writing. In addition, the lack of expertise in graphomotor skills and different pen grips also create high inter-person variability, resulting in significant disparities in the signals. The combination of paper surface and children's data compounds the difficulty
.

Designing age-adaptive models that generalize across adult and children writers on tablet is a highly challenging problem
, as it requires the model to handle heterogeneous writing styles within a single framework. 
Finally, achieving age-adaptive handwriting reconstruction on paper represents the most challenging configuration
. This combines pen-paper friction noise with extreme population diversity, which is currently unresolved. 

\begin{table}[ht]
    \centering
    \caption{Overview of state-of-the-art handwriting reconstruction approaches depending on the writing surface and target population. The difficulty is denoted by + signs and the more + signs, the higher the difficulty.}
    \label{tab:comparison}
    \resizebox{\textwidth}{!}{
    \begin{tabular}{|c|c|c|c|c|c|c|}
        \hline
        \multirow{2}{*}{\diagbox{\textbf{Surface}}{\textbf{Population}}} & \multicolumn{2}{c|}{\textbf{Adults}} & \multicolumn{2}{c|}{\textbf{Children}} & \multicolumn{2}{c|}{\textbf{Age-Adaptive}} \\
        \cline{2-7}
        & \textit{Difficulty} & \textit{Publication} & \textit{Difficulty} & \textit{Publication} & \textit{Difficulty} & \textit{Publication} \\
        \hline
        \textbf{Tablet} & + & \cite{Wehbi22}, \cite{Ott22}, \cite{Swaileh2023}, \cite{IMBERT2025PR} & ++++ & \cite{ADAPDA} & +++++ & Our\\
        \hline
        \textbf{Paper} & ++ & \cite{Imbert2025IGS} & +++++ & - & ++++++ & - \\
        \hline
    \end{tabular}
    }
\end{table}

Only a few works have addressed handwriting trace reconstruction from IMU signals. Wehbi et al.~\cite{Wehbi22} extend a preliminary study by \cite{Ott22}, initially conducted on a single writer, to a multi-writer setting. Their approach relies on deep learning techniques and uses handwriting collected on a tablet with an earlier version of the STABILO Digipen. The proposed Convolutional Neural Network reconstructs handwriting trajectories, with linear interpolation used to align sensor data with tablet-based ground truth for training.

More recently, \cite{Swaileh2023} propose a complete reconstruction pipeline based on the Digipen. This pipeline introduces a preprocessing stage using Dynamic Time Warping to align sensor signals with tablet-based ground truth, and a neural network inspired by Temporal Convolutional Networks (TCN). Building on this work, a mixture-of-experts approach has been proposed to account for different signal characteristics \cite{IMBERT2025PR}, with two expert models dedicated to pen-down and pen-up movements, respectively. All these experiments were conducted on datasets acquired on tablets and from adult handwriting.

In a recent work, handwriting reconstruction using the Digipen on paper was explored \cite{Imbert2025IGS}. Based on the mixture-of-experts approach discussed above, several strategies were investigated to handle paper-based writing, and a solution was proposed to embed the model into the Digipen. However, the quality of trajectory reconstruction on paper data remains lower than that obtained on tablet surfaces.

To the best of our knowledge, \cite{ADAPDA} is the only work investigating children’s handwriting using the Digipen. In this work, Imbert et al. propose a domain adaptation approach based on a Domain-Adversarial Neural Networks (DANN). Although the DANN model offers a good trade-off for using a single model to handle both adults and children handwriting, its performance falls short of that of source-specific dedicated models.

\section{Age-adaptive handwriting reconstruction using cross-domain learning}

Recent state-of-the-art methods achieve satisfactory handwriting trajectory reconstruction from the Digipen in controlled settings, typically involving adult writers producing short sequences. However, extending these approaches to more diverse populations, such as children, remains challenging \cite{ADAPDA}. As stated before, IMU signals vary significantly between adults and children due to differences in immature motor control and graphomotor development (cf. Section \ref{Dataset Overview and Analysis}). Children’s handwriting is generally less fluid, more variable in speed and pen holding, with abrupt movements resulting often in noisier and much longer signals, which directly impacts the quality of handwriting trace reconstruction.

These discrepancies in signal characteristics introduce a strong domain shift between adult and children data. As a result, a model trained exclusively on adult handwriting often fails to generalize effectively to children’s data, requiring dedicated adaptation strategies \cite{ADAPDA}.

\subsection{Motivation}


To address this limitation, we propose a cross-domain learning designed to improve robustness, modularity, and adaptability to heterogeneous handwriting styles. This framework builds upon the Temporal Convolutional Network (TCN) paradigm as such an architecture has been used with success for handwriting reconstruction from IMU signals \cite{IMBERT2025PR,Swaileh2023}. 

Joint training on adult and children data provides two key benefits. First, it increases data diversity, improving the model’s robustness to variations in writing dynamics and reducing sensitivity to noise and drift in IMU signals. Second, it allows the model to leverage complementary information: adult data provides more stable motion patterns, while children data introduces a wider range of variability, which is essential to capture for better generalization. 
By combining a shared representation with domain-specific heads, the proposed Dual-TCN architecture effectively captures both common and age-specific characteristics, leading to improved handwriting reconstruction across heterogeneous users.

In contrast to domain adaptation approaches such as \cite{ADAPDA}, where a performance trade-off is deliberately accepted in exchange for increased model genericity, our architecture follows a different design philosophy. Rather than sacrificing accuracy for universality, we concentrate the vast majority of the model's parameters in the shared encoder, which captures the common spatio-temporal structure of handwriting dynamics across all users. Only a small, lightweight set of domain-specific parameters, representing less than $5\%$ of the total parameter count, is allocated to the specialized heads. This asymmetric design ensures that the architecture remains suitable for embedded deployment, where memory and computational resources are constrained, while still preserving the ability to model population-specific characteristics with high fidelity.

\subsection{Dual-TCN: A two-head model based on TCN}

We introduce a novel neural architecture, called Dual-TCN, based on a two-head design to explicitly account for the domain shift between adult and children handwriting. Both types of data are jointly used during training, acting as a multi-domain learning strategy.

\begin{figure}[t]
\centering{
\includegraphics[width=0.99\textwidth]{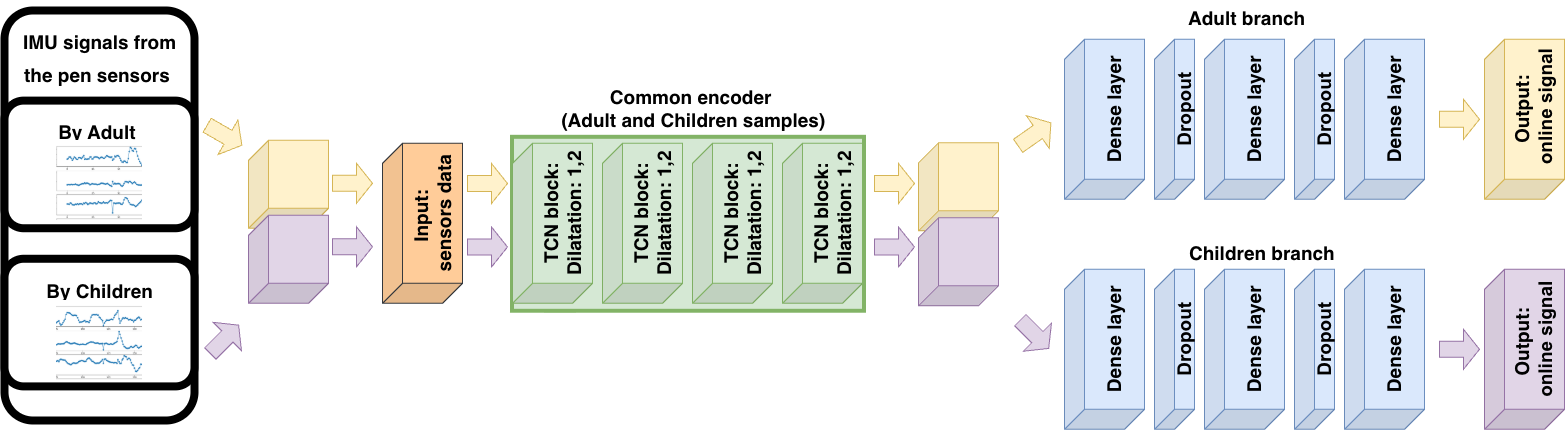}}
\caption{Dual‑TCN architecture. The model takes as input both adult and children writing signals in training, and one of the two at test time. A multivariate sequence based on IMU signals is given in input of the network and encoded by a shared TCN to produce high-level temporal features. Each branch consists of  fully connected layers that output the reconstructed handwriting trace for its corresponding domain (adults or children).
}
\label{fig:2branchs}
\end{figure}

As illustrated in Fig.~\ref{fig:2branchs}, the model relies on a shared TCN backbone that processes raw 7-channel IMU sequences and extracts high-level temporal representations common to both populations. TCN offer several practical advantages that make them particularly well-suited to this task. Unlike recurrent architectures such as LSTM, TCN support fully parallelized training, leading to significantly faster convergence \cite{bai2018empirical}. Their lightweight design also makes them well-adapted to training on datasets of limited size and to get a frugal model for embedding in the digital pen after further optimization. Furthermore, TCN capture local temporal context through dilated causal convolutions rather than encoding entire sequences globally, which aligns well with the local stroke dynamics inherent in handwriting motion. These properties, combined with their demonstrated strong performance in state-of-the-art sequential modeling approaches and broad adoption across various domains \cite{ICTAI}, make TCN a particularly suitable backbone for handwriting trajectory reconstruction from inertial data.

On top of this shared encoder, two parallel branches are introduced as domain-specific heads:
\begin{itemize}
\item \textbf{Adult branch}, optimized for more stable and fluent handwriting signals, typically characterized by smoother dynamics and lower variability.
\item \textbf{Children branch}, designed to handle noisier and more irregular signals, reflecting variability in motor control and graphomotor development.
\end{itemize}

Each branch consists of a sequence of fully connected layers and output the final coordinate predictions. This multi-head design enables the model to learn shared temporal dynamics while specializing the decoding process for each population.

The architecture is summarized in Table~\ref{tab:dual_tcn_architecture}. The shared TCN backbone is composed of four stack or temporal convolutional blocks with 256 channels, kernel size 3, and a dilation pattern following $[1, 2]$ repeated across blocks. Each convolutional layer is followed by a ReLU activation and a dropout rate of 0.2. On top of the backbone, two parallel domain-specific branches (Adult and Children) are defined. Each branch consists of three fully connected layers that progressively reduce the feature dimensionality. ReLU activations and dropout are applied after the first two linear layers, while the final layer outputs the $(x, y)$ trajectory coordinates.

The model totals $3,047,300$ parameters, distributed across a shared backbone and two lightweight domain-specific heads. The shared TCN backbone accounts for the overwhelming majority of the capacity, with $2,964,736$ parameters ($97.3\% $), reflecting the assumption that low-level temporal feature extraction is domain-agnostic and therefore benefits from joint training across both populations. Each domain-specific branch contributes  $41,282$ parameters ($1.35\% $), forming a  compact head designed to adapt the shared representation to the statistical characteristics of its target population, whether adult or children.

\setlength{\tabcolsep}{8pt}
\renewcommand{\arraystretch}{1.}

\begin{table}[t]
\centering
\caption{Detailed architecture of the Dual-TCN model.}
\label{tab:dual_tcn_architecture}
\resizebox{\columnwidth}{!}{
\begin{tabular}{c c c c c c c}
\hline
\textbf{Part} & \textbf{Layer} & \textbf{In} & \textbf{Out} & \textbf{Layer} & \textbf{dilation} & \textbf{dropout} \\
\hline

\multirow[c]{8}{*}{TCN Backbone}
 & \multirow{2}{*}{TCN Block} & 7   & 256 & \multirow{3}{*}{2} & 1 & \multirow{3}{*}{0.2} \\
 &                            & 256 & 256 &                    & 2 & \\
\cline{2-7}
 & \multirow{2}{*}{TCN Block} & 256 & 256 & \multirow{3}{*}{2} & 1 & \multirow{3}{*}{0.2} \\
 &                            & 256 & 256 &                    & 2 & \\
\cline{2-7}
 & \multirow{2}{*}{TCN Block} & 256 & 256 & \multirow{3}{*}{2} & 1 & \multirow{2}{*}{0.2} \\
 &                            & 256 & 256 &                    & 2 & \\
\cline{2-7}
 & \multirow{2}{*}{TCN Block} & 256 & 256 & \multirow{3}{*}{2} & 1 & \multirow{2}{*}{0.2} \\
 &                            & 256 & 256 &                    & 2 & \\
\hline


\textbf{Part} & \multicolumn{3}{c}{\textbf{Adult}} 
 & \multicolumn{3}{c}{\textbf{Children}} \\
\cline{1-7}

 & \textbf{Layer} & \textbf{In} & \textbf{Out} 
 & \textbf{Layer} & \textbf{In} & \textbf{Out} \\
\cline{2-7}

\multirow[c]{7}{*}{Specific heads} & Linear 1  & 256 & 128 & Linear 1  & 256 & 128 \\
 & ReLU      &     &     & ReLU      &     &     \\
 & Dropout 0.2   &     &     & Dropout 0.2  &     &     \\

 & Linear 2  & 128 & 64  & Linear 2  & 128 & 64  \\
 & ReLU      &     &     & ReLU      &     &     \\
 & Dropout 0.2  &     &     & Dropout 0.2   &     &     \\

 & Linear 3  & 64  & 2   & Linear 3  & 64  & 2   \\

\hline
\end{tabular}
}
\end{table}

\subsection{Training}

The model takes IMU signals acquired from the pen sensors as input and predicts the associated online handwriting trajectories, represented as sequences of displacement vectors $(\Delta x, \Delta y)$. The training objective is defined as the Mean Squared Error (MSE) between the predicted and ground-truth displacements. Optimization is performed using the Adam algorithm, with an early stopping strategy (patience $= 100$) to prevent overfitting.

During training, each batch is constructed to include an equal number of adult and children samples in order to ensure balanced learning across both populations. Specifically, we leverage the force sensor to split the IMU signals into touching strokes, which are then provided to the model as input. All input sequences are first processed by the shared TCN encoder, which extracts latent temporal representations. These shared features are then routed to the corresponding domain-specific head: the adult branch receives features from adult samples, while the children branch processes those from children samples.

Each branch is thus optimized on its respective data distribution, enabling specialization to age-specific handwriting dynamics while still leveraging shared representations learned across both domains. This training strategy promotes robustness to signal variability and enhances the model’s ability to generalize across different levels of graphomotor development.

\section{Dataset Overview and Preliminary Analysis}
\label{Dataset Overview and Analysis}

\subsection{Dataset Description}

The experimental dataset comprises two distinct populations recorded under the same acquisition setup. Data were collected using a custom instrumented digital pen embedding IMU sensors (accelerometer, gyroscope and force sensor), mechanically coupled with a Wacom electromagnetic resonance (EMR) stylus inserted into the pen body.
The pen was used on a tablet that simultaneously captures:
\begin{itemize}
    \item the inertial signals tri-axial accelerometer and gyroscope and force sensor via the embedded IMU
    \item the online trajectory of the stylus tip position, pressure, and hover distance via the EMR digitizer
\end{itemize}

The corpus is partitioned into three non-overlapping splits train, validation, and test whose sizes are reported in Table~\ref{tab:dataset_recap}\footnote{The anonymized dataset is available for research purposes at the following link X.}.

\begin{table}[h]
\centering
\caption{Dataset split summary: number of sequences and touch strokes
         per population and per split. A sequence refers to a complete handwriting trajectory, consisting of both touch strokes and pen-up events. }
\label{tab:dataset_recap}
\renewcommand{\arraystretch}{1.25}
\begin{tabular}{llrr}
\hline
\textbf{Population} & \textbf{Split} & \textbf{Sequences} & \textbf{Touch strokes} \\
\hline
Adult  & Train      & 5{,}189 & 24{,}297 \\
Children  & Train      & 1{,}085 &  4{,}822 \\
\hline
Adult  & Validation & 1{,}889 &  8{,}722 \\
Children  & Validation &   222   &    917   \\
\hline
Adult  & Test       &   272   &  1{,}212 \\
Children  & Test       &   156   &    635   \\
\hline
\end{tabular}
\end{table}

As visible from Table~\ref{tab:dataset_recap}, the adult split is substantially larger than the child split across all partitions. This imbalance reflects the inherent practical difficulty of collecting data from minors in classrooms: each recording session requires prior  institutional approval, parental informed consent, and teacher scheduling coordination, all of which significantly constrain the volume and throughput of child data acquisition compared to adult volunteers. 

\subsection{Comparison of Graphomotor Descriptors}

\begin{figure}[ht]
\centerline{\includegraphics[width=0.99\textwidth]{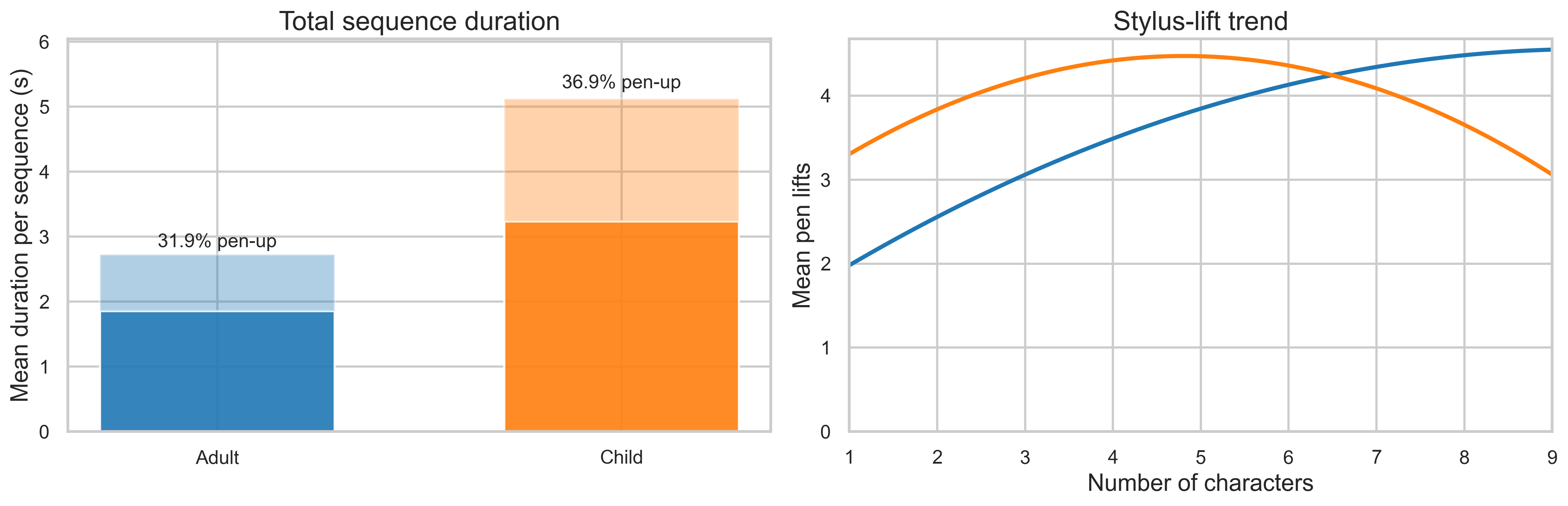}}
\caption{Pen-state duration and stylus-lift behavior in adults and children.
Left: mean sequence duration for each group, decomposed into pen-down (solid color) and pen-up (transparent color) time. Right: stylus-lift trend as a function of characters length, with adults in blue and children in orange.}
\label{fig:Graphomotor}
\end{figure}

Figure~\ref{fig:Graphomotor} presents pen-state timing and stylus-lift behaviour across the two age-groups.

\begin{figure}[ht]
\centerline{\includegraphics[width=0.99\textwidth]{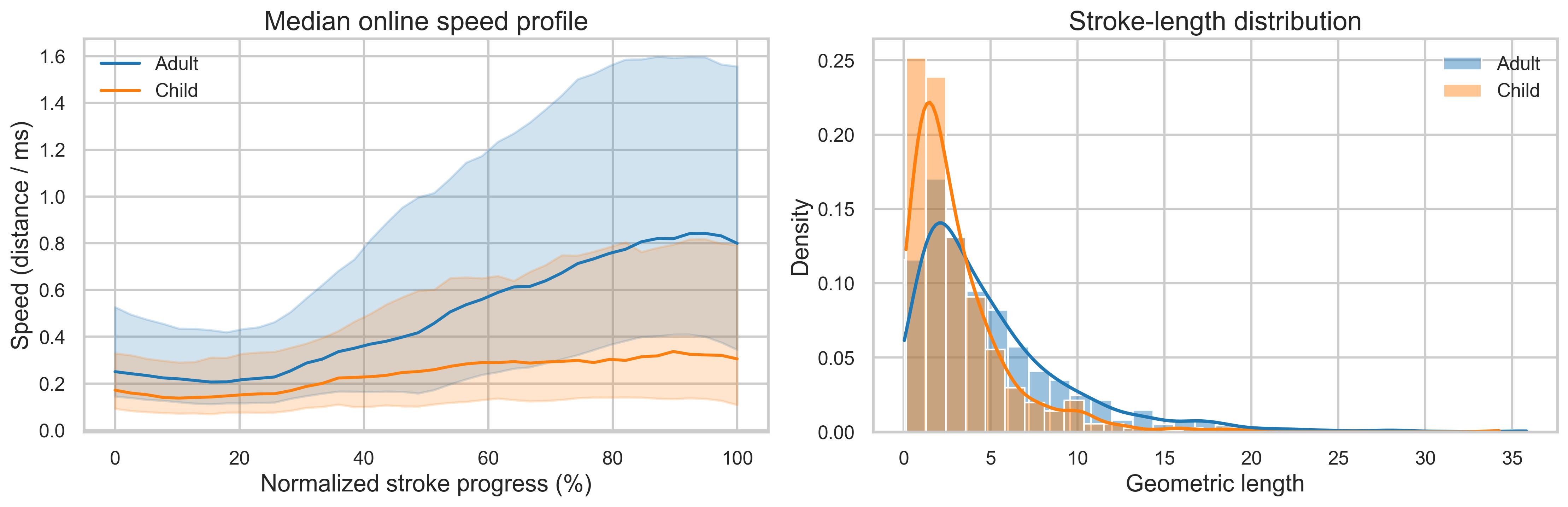}}
\caption{Left: median online speed profiles for adults (blue) and
         children (orange) as a function of normalized stroke progress; shaded bands denote the inter-quartile range. Adults display a late-peaking, anticipatory velocity
         profile whereas children exhibit a flat, low-amplitude curve.
         Right: stroke length distribution.}
\label{fig:Speed_Curves}
\end{figure}


Children exhibit substantially longer sequence durations than adults, close to double that of adults, indicating markedly less efficient overall writing execution. In both groups a considerable proportion of sequence time is spent with the pen lifted from the surface, suggesting that pen-up pauses constitute a structurally important component of the handwriting process regardless of age. However, pen-up are slightly more pronounced for children ($36.9\%$ of the time vs $31.9\%$ for adults).
The stylus-lift trend as a function of sequence length reveals a qualitative divergence between groups. Adults show a monotonically increasing number of pen lifts with character count, consistent with a stable, letter-by-letter lifting strategy that scales predictably with word length. Children, by contrast, display an inverted pattern, peaking around 5 characters before declining. This behaviour suggests that they lift the stylus more frequently for shorter sequences due to a lack of confidence, yet as sequence length grows, the natural continuity of cursive writing takes over, reducing the number of pen lifts and explaining the observed decline. Together, these findings point to more automatised and temporally efficient pen-state management in adults compared with children.

To capture the temporal dynamics of graphomotor execution, Figure~\ref{fig:Speed_Curves} displays the median online speed profile of each group, computed by interpolating instantaneous speed onto a 
normalised stroke-progress axis. Adults consistently achieve higher speeds at every point along the normalised stroke trajectory.
More informatively, the shape of the adult profile reveals a progressive velocity build-up that peaks late in the stroke before a rapid deceleration, a hallmark of anticipatory motor control.
Children, by contrast, display a flat, low-amplitude profile throughout the stroke, suggesting that execution speed is only weakly modulated across the gesture. This pattern is consistent with an online, feedback-driven control mode rather than the predictive, open-loop execution predominant in expert writers. The mean speed in Figure~\ref{fig:Speed_Curves} confirms these
observations at the stroke level, showing that the entire speed distribution of children is shifted downward relative to adults.  It is worth noting that adults contribute a larger overall number of strokes, resulting in higher density across most of the distribution. However, in the specific case of short strokes, children exhibit a higher density, reflecting their more frequent pen lifts, which segment the trajectory into a greater number of shorter, fragmented strokes compared to the more fluid and continuous writing of adults.

Taken together, the graphomotor and speed-profile analyses demonstrate clear, statistically strong differences in both the magnitude and the temporal structure of motor execution between adults and children, providing well-founded empirical justification for training age-specific graphomotor heads within the Dual-TCN framework.

\section{Results}

Model performance is evaluated following the evaluation protocol introduced by \cite{Swaileh2023}, which consists of computing the Fréchet distance ($\downarrow$) between the reconstructed handwriting trajectory and the ground-truth trace, where lower values indicate higher reconstruction fidelity. To account for variability in training, each model is trained and evaluated over three independent runs, and the mean and standard deviation of the performance metrics are reported.

\begin{figure}[ht]
\centerline{\includegraphics[width=0.99\textwidth]{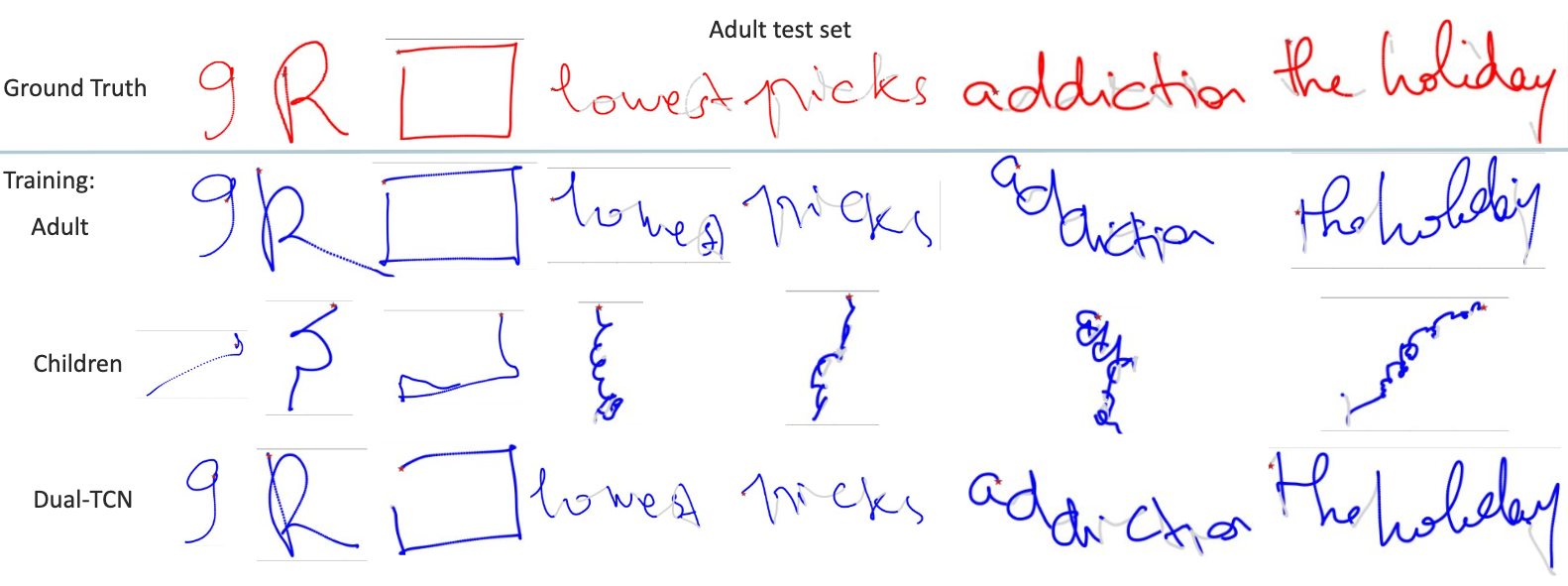}}
\caption{Examples of ground truth traces and reconstructed traces from adult test data produced by the state-of-the-art model from \cite{Swaileh2023}) trained on adult data only (first blue line), on children data only (second blue line) and by our Dual-TCN model (last blue line).}
\label{fig:results_adults}
\end{figure}

\begin{figure}[ht]
\centerline{\includegraphics[width=0.99\textwidth]{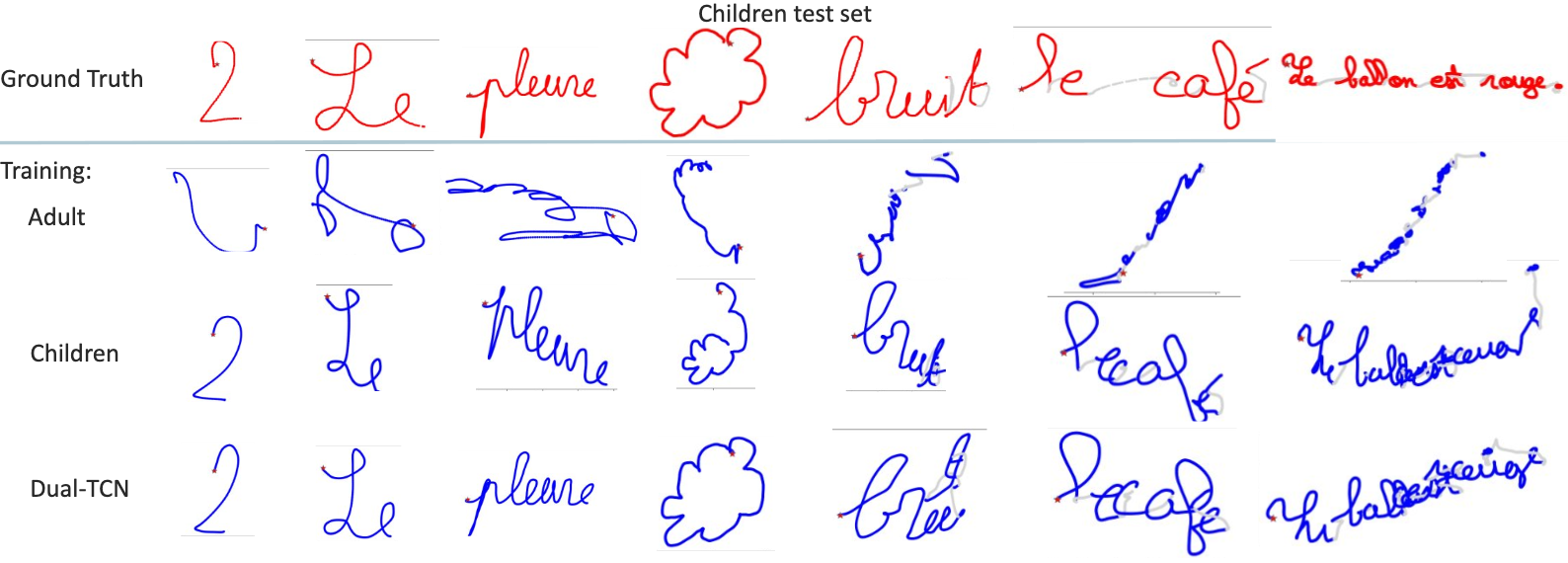}}
\caption{Examples of ground truth traces and reconstructed traces from children test data produced by the state-of-the-art model from \cite{Swaileh2023}) trained on adult data only (first blue line), on children data only (second blue line) and by our Dual-TCN model (last blue line).}
\label{fig:results_children}
\end{figure}

We reimplemented the approach of \cite{Swaileh2023} and \cite{IMBERT2025PR} then trained it from scratch on two distinct training sets: the Adult dataset and the Children dataset. Quantitative and qualitative results are reported in Table~\ref{tab:sota} and Fig.~\ref{fig:results_adults} and~\ref{fig:results_children} respectively. All methods are evaluated on both adult and children test sets, and their performance is compared with our proposed Dual-TCN model. Although the DANN model in \cite{ADAPDA} address the same problem of handling adults and children handwriting, a direct comparison is not viable for two reasons. First, \cite{ADAPDA} relies on a different evaluation methodology, making numerical comparison uninformative. Second, while the DANN model proposed in the paper offers a good trade-off by using one model to process data from both adults and children, it does not outperform the two specific models tailored to individual sources (adults or children), whose network architecture is derived from~\cite{Swaileh2023}. Hence, we compare our work to~\cite{Swaileh2023} and~\cite{IMBERT2025PR}. 

\begin{table}[ht]
    \centering
    \caption{Mean Fréchet distance ($\downarrow$) between reconstructed trajectory and ground truth. Our proposed Dual-TCN is compared to the main state-of-the-art approaches \cite{Swaileh2023} and \cite{IMBERT2025PR} trained on adult-only and children-only stroke data and evaluated at sequence level.}
    \label{tab:sota}
    \resizebox{\columnwidth}{!}{%
    \begin{tabular}{|c|c|c|c|c||c|}
        \hline
        Model & \multicolumn{2}{c|}{TCN \cite{Swaileh2023}} & \multicolumn{2}{c||}{MOE \cite{IMBERT2025PR}} & our Dual-TCN \\
        \hline
        Training set & Adult & Children & Adult & Children & Adult + Children \\
        \hline
        Adult test & $0.346 \pm 0.004$ & $0.785 \pm 0.002$ & $\textbf{0.336} \pm 0.014$ & $0.724 \pm 0.009$ & $\textbf{0.336} \pm 0.014$ \\
        \hline
        Children test & $0.674 \pm 0.030$ & $0.448 \pm 0.050$ & $0.674 \pm 0.013$ & $0.429 \pm 0.032$ & $\textbf{0.425} \pm 0.021$ \\
        \hline
    \end{tabular}}
\end{table}

The results demonstrate the effectiveness of the proposed Dual-TCN architecture across both populations. On the adult test set, Dual-TCN achieves the lowest Fréchet distance ($0.336$), outperforming \cite{Swaileh2023} and \cite{IMBERT2025PR} trained on adult data alone ($0.346$ and $0.336$), despite the latter benefiting from a homogeneous, in-domain training set. Especially, our Dual-TCN has nearly half the parameters of the state-of-the-art MoE model~\cite{IMBERT2025PR}, as the latter combines two expert models similar to the TCN from \cite{Swaileh2023}. More notably, on the children test set, Dual-TCN yields a mean Fréchet distance of $0.425$, outperforming both the adult-trained baseline and the children-trained baseline, while being trained on the combined dataset. To our knowledge, this is the first work that successfully combines both datasets in a joint learning framework, since in~\cite{ADAPDA} the authors demonstrated that the combination is a good trade-off but performs worse than task-specific models. This result highlights the ability of the multi-head design to effectively disentangle age-specific writing dynamics without sacrificing performance on either domain. Moreover, beyond raw accuracy, Dual-TCN stay consistent across runs, indicating stable optimization and robustness to random initialization. 

Qualitative results in Fig.~\ref{fig:results_adults} for adults and Fig.~\ref{fig:results_children} for children further confirm the model's robustness across all domains, with Dual-TCN producing reconstructed traces that more faithfully capture the graphomotor patterns of both adult and children handwriting, reflecting the richer variability learned from the combined training set. In particular, touch strokes are significantly better reconstructed using our Dual-TCN. However, next stroke repositioning after pen-up events is sometimes unsatisfactory, despite improvements, mainly because the model was trained only on touch strokes. We opted for this training to improve handwriting reconstruction, as pen-up event management remains a challenging and open problem~\cite{IMBERT2025PR}. 

\section{Conclusion}

In this work, we proposed a cross-domain learning for handwriting trajectory reconstruction from IMU signals acquired with a digital pen. We designed a novel multi-head neural network architecture, called Dual-TCN, combining a shared TCN backbone with age-specific decoding heads. In this way, the model jointly leverages adult and children data during training, effectively addressing the domain shift induced by differences in graphomotor development while benefiting from data in both domains.

Experimental results demonstrate that our Dual-TCN achieves state-of-the-art reconstruction quality on both adult and children test sets, outperforming single-population baselines while exhibiting improved training stability. Crucially, the architecture is designed with embedded deployment in mind: since the two branches diverge only at the final layer, the memory overhead is minimal making the model directly compatible with integration into the pen hardware.

Future work will focus on two main directions. First, we will explore further compression strategies to optimize the model for real-time embedded inference. Second, a dedicated effort will focus on improving pen-up segments  repositioning, which remains the most challenging aspect of trajectory reconstruction. During pen-up events, the absence of tip-force signal and the 
unconstrained nature of the motion lead to significant drift. Thus, designing specific architectural or training strategies to better handle these segments represents a key way to improve overall handwriting reconstruction quality.

\begin{credits}
\subsubsection{\ackname} This work was supported by the Kempestiftelserna grant JCSMK23-0109. 
The computations were enabled by the Berzelius resource provided by the Knut and Alice Wallenberg Foundation at the National Supercomputer Centre.
\end{credits}

%
%
%
\bibliographystyle{splncs04}
\bibliography{bib}

\end{document}